\documentclass[10pt,twocolumn,letterpaper]{article}

\usepackage[pagenumbers]{cvpr} 
\usepackage{multirow}

\usepackage{placeins} 
\usepackage{colortbl} 
\usepackage{xcolor}   
\usepackage{float}  
\usepackage{booktabs}  

\definecolor{cvprblue}{rgb}{0.21,0.49,0.74}
\usepackage[pagebackref,breaklinks,colorlinks,allcolors=cvprblue]{hyperref}

\def\paperID{57}
\def\confName{EarthVision}
\def\confYear{2025}

\title{SAR-to-RGB Translation with Latent Diffusion for Earth Observation}

\author{
Kaan Aydin \\
University of St. Gallen \\
{\tt\small kaan.aydin@student.unisg.ch}
\and
Joëlle Hanna \\
University of St. Gallen \\
{\tt\small joelle.hanna@unisg.ch}
\and
Damian Borth \\
University of St. Gallen \\
{\tt\small damian.borth@unisg.ch}
}

\begin{document}
\maketitle
\begin{abstract}
Earth observation satellites like Sentinel-1 (S1) and Sentinel-2 (S2) provide complementary remote sensing (RS) data, but S2 images are often unavailable due to cloud cover or data gaps. To address this, we propose a diffusion model (DM)-based approach for SAR-to-RGB translation, generating synthetic optical images from SAR inputs. We explore three different setups: two using Standard Diffusion, which reconstruct S2 images by adding and removing noise (one without and one with class conditioning), and one using Cold Diffusion, which blends S2 with S1 before removing the SAR signal. We evaluate the generated images in downstream tasks, including land cover classification and cloud removal. While generated images may not perfectly replicate real S2 data, they still provide valuable information. Our results show that class conditioning improves classification accuracy, while cloud removal performance remains competitive despite our approach not being optimized for it. Interestingly, despite exhibiting lower perceptual quality, the Cold Diffusion setup performs well in land cover classification, suggesting that traditional quantitative evaluation metrics may not fully reflect the practical utility of generated images. Our findings highlight the potential of DMs for SAR-to-RGB translation in RS applications where RGB images are missing.
\end{abstract}

\section{Introduction}
\label{sec:intro}

Earth observation plays an important role in environmental monitoring \cite{Hanna2023PhysicsGuidedML}, disaster management \cite{Adriano2020LearningFM}, and land-use analysis \cite{ThomasRamos2024MultispectralSS, Hanna2023SparseMV, Scheibenreif2022SelfsupervisedVT}. The Sentinel-1 (S1) \cite{torres2012gmes} and Sentinel-2 (S2) \cite{drusch2012sentinel} missions, part of the European Space Agency's Copernicus program, provide complementary remote sensing (RS) data. S1 uses Synthetic Aperture Radar (SAR) to capture detailed images of Earth's surface, regardless of weather conditions and time of day. In contrast, S2 employs a multispectral instrument to capture optical images across 13 spectral bands.

Combining these two modalities enables comprehensive surface monitoring, improving applications such as vegetation analysis, change detection, and disaster response \cite{haftner2022, podromou2023, Notarnicola2017}. However, S2 images are often unavailable or difficult to use due to cloud cover, bad weather, or gaps in data collection, resulting in missing or incomplete information. This lack of paired S1-S2 observations poses challenges for multimodal applications that rely on both data sources.

SAR-to-optical image translation aims to bridge this gap by generating S2 images from S1. This task is inherently challenging due to the fundamental differences between SAR and optical image modalities. SAR captures surface structure and texture but lacks spectral reflectance information, making the direct translation difficult. In this work, we focus on translating SAR images to the RGB bands of S2, as they are the most widely used for visual interpretation and downstream applications. Traditionally, Generative Adversarial Networks (GANs) \cite{Goodfellow2014GenerativeAN} have been used for SAR-to-RGB translation, but they suffer from mode collapse and instability during training. To overcome these limitations, we propose a Diffusion Model (DM) for SAR-to-RGB translation to generate high-fidelity optical images. Compared to GANs, DMs offer more stable training and better preservation of structural details \cite{Dhariwal2021DiffusionMB}.

Our approach consists of three steps. First, we train a DM to learn the mapping from S1 to S2 images. Second, we generate synthetic S2 images from S1 inputs. Third, we evaluate the generated images in practical downstream tasks, such as land cover classification and cloud removal, to assess both their visual quality and usability in RS applications. In summary, the contributions of our work are as follows: 

\begin{enumerate}
    \item We analyze different diffusion strategies (Standard vs. Cold Diffusion) and conditioning strategies (with and without class conditioning) for SAR-to-RGB translation, exploring how to best leverage DMs for RS applications.
    \item We systematically evaluate how well the synthetically generated RGB images perform in real-world RS tasks, such as land cover classification and cloud removal, beyond just their visual quality.
\end{enumerate}

\section{Related Works}
\label{sec:related_works}

\subsection{SAR-to-RGB translation}
SAR-to-RGB translation is a specialized task within the broader field of Image-to-Image (I2I) translation that aims to convert SAR imagery into visible RGB images.

Recent advances in deep learning, particularly GANs \cite{Goodfellow2014GenerativeAN}, have shown promising results in this domain. For example, Fu \textit{et al}. \cite{fu2019} proposed a GAN-based approach for translating SAR images into optical images, utilizing multiscale discriminators to improve the quality of the generated images. Despite its advancements, this method faces challenges such as producing blurred boundaries and failing to accurately capture the geometric details of certain objects, particularly buildings. 

The remarkable success of DMs in image translation has led to a growing body of research focused on leveraging their capabilities. For instance, Bai \textit{et al}. \cite{bai2024_first} proposed a conditional diffusion approach for translating SAR images to RGB, utilizing a U-Net \cite{Ronneberger2015UNetCN} architecture as the backbone.

In parallel, more sophisticated approaches have explored broader applications of DMs. Berian \textit{et al}. \cite{berian2025} introduced \textit{CrossModalityDiffusion}, a framework designed for multi-modal novel view synthesis. This method incorporates modality-specific encoders alongside DMs to generate consistent outputs across diverse modalities. Although not specifically focused on SAR-to-RGB translation, this approach highlights the broader potential of DMs in facilitating image translation across a wide range of modalities. 

\subsection{Applications of DMs in RS}
Since their introduction to RS in 2021, DMs have gained significant traction, demonstrating great potential in advancing image processing and analysis for RS \cite{Liu_2024}. Beyond SAR-to-RGB translation, DMs have impacted various other areas of RS. 

For example, Khanna \textit{et al.} \cite{khanna2024} integrated additional numerical metadata, such as geolocation and sampling time, as prompts for a Stable Diffusion (SD) model. This approach enriched the model’s input, improving its ability to generate high-quality satellite images.

Super-resolution is another domain benefiting from DMs. Han \textit{et al.} \cite{han2023} combined Transformer and CNN architectures to extract global and local features from multispectral low-resolution images. These fused feature representations were then used to guide a DM in generating high-resolution images in a single training step.

Cloud removal is also a critical challenge in RS, as optical images are frequently obstructed by clouds. Zou \textit{et al.} \cite{zou2023} addressed this issue by developing DiffCR, a method that first extracts features from the cloudy image and its noise level. These extracted spatial and temporal features are then incorporated into the DM as control conditions, enhancing its ability to reconstruct cloud-free images.

\subsection{Advancements in DMs}
DMs have emerged as the state-of-the-art approach for generative modeling and have overtaken GANs due to their ability to produce high-quality output with better diversity and a more stable training regime. DMs were first introduced by Sohl-Dickstein \textit{et al.} \cite{sohl2015} and started to garner more attention following the publication of the Denoising Diffusion Probabilistic Model (DDPM) by Ho \textit{et al.} \cite{ho2020} who formalized the framework as a generative process based on a forward diffusion process that gradually adds noise to data and a learned reverse process that denoises it step by step. Subsequent works introduced innovations that span multiple directions.  

Rombach \textit{et al}. \cite{rombach2022} introduced a novel approach \textbf{Latent Diffusion Models (LDMs)} for scalable high-dimensional image generation. Instead of directly operating on raw pixel data, LDMs first encode high-dimensional images into a lower-dimensional latent representation using a Variational Autoencoder (VAE). By performing the computationally heavy diffusion step in the latent space rather than in the pixel space, one can reduce memory and computational requirements, leading to faster training and inference. Also, LDMs provide a simplified learning task as compressing the original space into a latent space helps the model focus on the most important details.

It is common to use conditioning information to steer the image generation process towards a desired outcome. The conditioning information can be in the form of class labels, semantic maps, or a descriptive prompt. One widely discussed approach is \textbf{Classifier-Free Guidance} \cite{ho2022}, in which the DM is trained with two modes (conditional and unconditional mode) by removing the conditioning for a certain fraction of time (e.g., 10\%). During inference, the unconditional and conditional predictions are blended together to produce the final output. 

In their paper, Bansal \textit{et al.} \cite{bansal2022} introduce \textbf{Cold Diffusion}, a framework that re-examines the need for Gaussian noise, or any randomness at all, for DMs to work in practice. They reconsider DMs built around arbitrary image transformations, such as blurring or downsampling. In one of their experiments, they include a new transformation called \textit{Animorphosis}, in which they iteratively transformed a human face from CelebA to an animal from AFHQ. However, in principle, they argue that such an interpolation can be done for any two initial data distributions. 
\begin{figure*}[ht]
    \centering
    \includegraphics[width=0.9\textwidth]{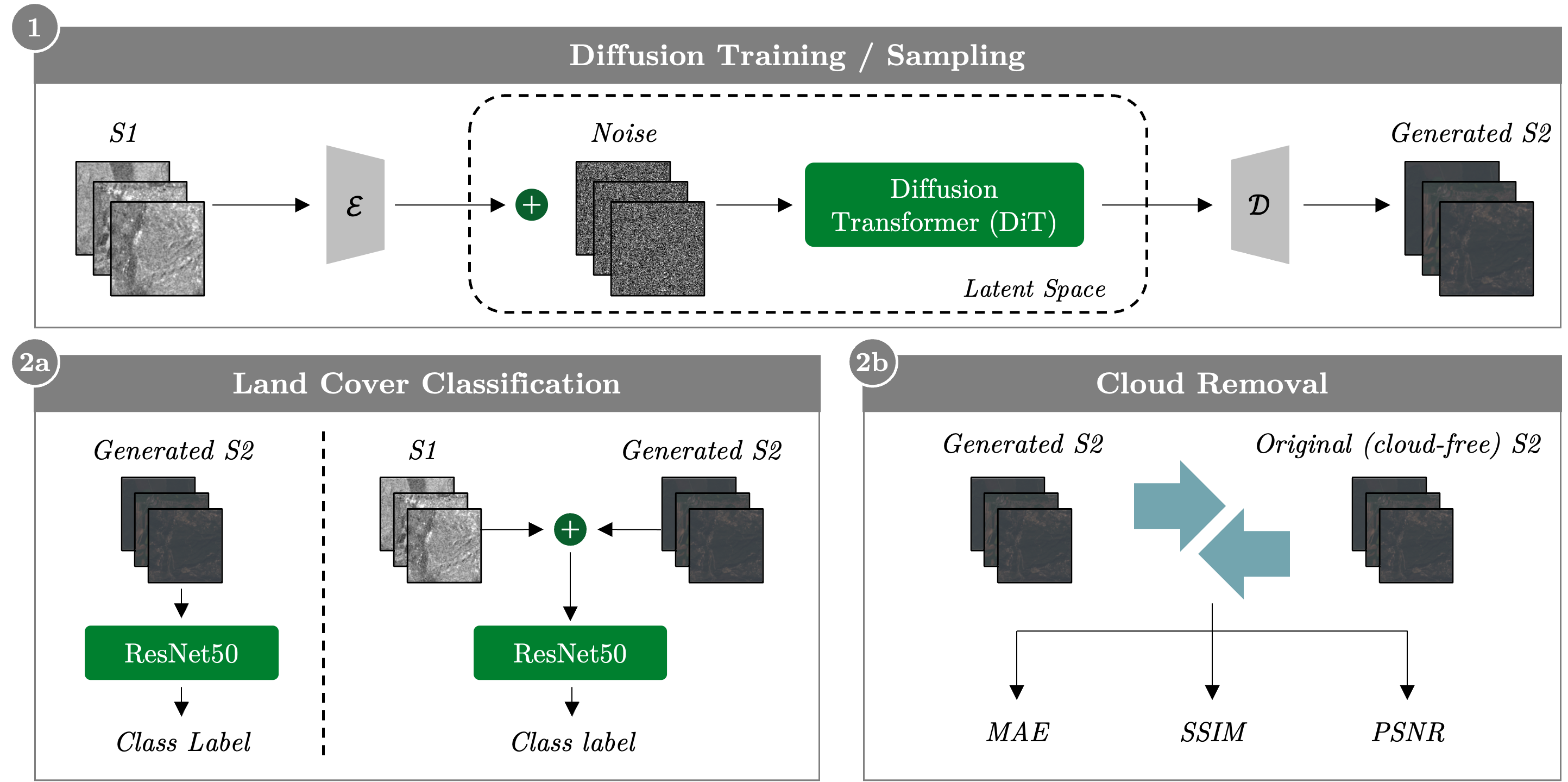}
    \caption{Schematic overview of our standard diffusion methodology. In the first step, Diffusion Training \& Sampling, we train our diffusion models that take a SAR image as input and predict the corresponding RGB output. Once training is complete, we freeze the DMs and use them to generate synthetic RGB images (Generated S2). In the next step (2a and 2b), these generated samples are then used for evaluating our downstream applications, specficially Land Cover Classification and Cloud Removal. The details of our setup are discussed in the following sections.}
    \label{fig:methodology}
\end{figure*}

\section{Methodology}
\label{sec:methdology}
Our research aims to generate synthetic RGB images from SAR data and evaluate their utility in downstream tasks. Our approach builds upon the method introduced by Bai \textit{et al}. \cite{bai2024_first}. However, rather than employing a U-Net, we adopt the Vision Transformer (ViT) \cite{dosovitskiy2021} as the backbone for our DM, leveraging its strengths in capturing global context and complex dependencies. Additionally, we perform diffusion in the latent space by encoding images with a Variational Autoencoder (VAE), reducing computational complexity while preserving essential features. To further evaluate the performance of our method, we extend our approach by incorporating two downstream tasks as additional evaluation benchmarks. Figure \ref{fig:methodology} shows the schematic overview of our methodology.

\subsection{Diffusion Training}
In this section, we describe the training process of our Standard Diffusion approach and will highlight the differences with the Cold Diffusion method in Section \ref{sec:cold_diffusion}.

\subsubsection{Data Preparation}
\label{sec:data_preparation}
Our setup uses paired SAR (from S1) and RGB (from S2) images as input. Following Bai \textit{et al.} \cite{bai2024_first}, we use only the VV channel from S1 and replicate it across three channels. As commonly done, we clip the SAR values to the range $\left[ -25, 0 \right]$ and the RGB values to $[0, 10000]$. To normalize the data, we divide the SAR values by 25 and the RGB values by $10000$, scaling them to the range $\left[ 0, 1 \right]$.  Finally, we standardize the normalized images to have a mean of 0.5 and a standard deviation of 0.5, as required by the VAE input.

\subsubsection{Model Architecture}
\label{sec:model_architecture}
Our approach follows the architecture proposed by Peebles \& Xie \cite{peebles2023}. Our diffusion framework consists of (i) a VAE that compresses images into a latent space and (ii) a DM that operates in this latent space, focusing on high-level semantics for better generalization and synthesis. For the VAE, we use a pre-trained Exponentially Moving Average (EMA) model from Stable Diffusion \cite{stabilityai_sd_vae}. For the DM, we adopt the Diffusion Transformer (DiT) proposed by Peebles \& Xie \cite{peebles2023}, which replaces the traditional U-Net with a ViT and substitutes the cross-attention mechanism with adaptive layer normalization (AdaLN). We use the ``DiT-S/4'' architecture, which has 12 layers, 6 heads, a latent size of 384, and a patch size of 4.

The \textbf{forward pass} is described as follows: First, both SAR and RGB images ($256 \times 256 \times 3$)\footnote{For SAR, we replicate the VV channel across three channels} are processed through the VAE encoder $\mathcal{E}$, which compresses them into a latent representation ($32 \times 32 \times 4$), given the VAE encoder $\mathcal{E}$'s downsampling factor of 8. The DiT receives an input of size $32 \times 32 \times 8$, formed by concatenating a noise vector of size $32 \times 32 \times 4$ with the SAR latent representation of the same size. The SAR data is concatenated to serve as conditioning, guiding the image generation process to align with the semantic characteristics of the SAR representation \cite{bai2024_first}. This differs from the original setup by Peebles \& Xie \cite{peebles2023}, which only uses a noise vector. The input is patchified into a sequence of tokens $T$ and then processed by subsequent DiT blocks. The final DiT block produces a sequence of tokens that are decoded into two outputs: a noise and variance prediction. Finally, after sampling from the DM, the VAE decoder $\mathcal{D}$ reconstructs the final output in pixel space as $x = \mathcal{D}(z)$. 

\paragraph{Auxiliary Information:} Before passing the tokens into the DiT blocks, we incorporate auxiliary information such as noise timesteps $t$ and class labels $c$. The timesteps are encoded using sinusoidal embeddings, while 
 class labels are processed through an embedding layer. These embeddings are then summed and combined with the tokens before being fed into the DiT blocks.

\subsubsection{Training Objective}
During training, a random timestep is sampled for each input, and the model learns to predict the corresponding noise. Our DM is trained using a Mean Squared Error (MSE) loss, which minimizes the difference between the predicted and actual noise at that timestep. In addition to predicting the mean, our model also estimates variance dynamically for adaptive noise removal, rather than using a fixed value. We incorporate a Variational Lower Bound (VLB) loss to align the predicted variance with the true noise distribution. The final training objective combines both losses to ensure stable training and accurate predictions of both the mean and variance. The complete objective can be expressed as:

\begin{equation}
L_{final} = L_{MSE} + L_{VLB}
\end{equation}

\subsubsection{Cold Diffusion}
\label{sec:cold_diffusion}
\begin{figure}[ht]
    \centering
    \includegraphics[width=0.45\textwidth]{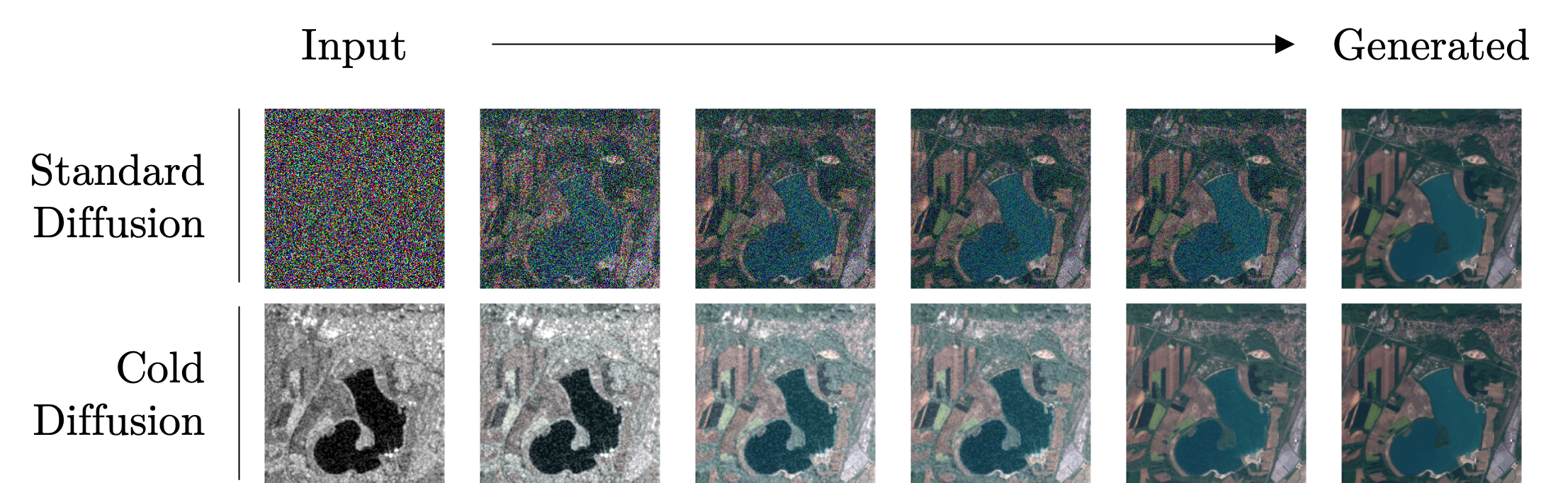}
    \caption{Visualisation of the generation process for Standard Diffusion (top row) and Cold Diffusion (bottom row)}
    \label{fig:process_visualisation}
\end{figure}

In addition to our Standard Diffusion setup, we explore an alternative setup known as Cold Diffusion, inspired by Bansal et al. \cite{bansal2022}. Unlike traditional diffusion models, which gradually add Gaussian noise to the target image and learn to reverse the process, Cold Diffusion follows a different approach: instead of adding random noise, it progressively blends the SAR image with the RGB image during training. The model then learns to remove the SAR signal step by step to reconstruct the RGB image (Figure \ref{fig:process_visualisation}). Formally, we state: 

\begin{equation}
x_t = \sqrt{\alpha_t}x + \sqrt{1-\alpha_t}z
\label{eq:cd_noisification}
\end{equation}

where \( x \) represents the target (RGB) distribution and \( z \) the source (SAR) distribution.

To sample from the learned distribution, a SAR image is drawn from the source distribution, and the transformation is reversed. Bansal \textit{et al.} \cite{bansal2022} observed that the original sampling algorithm performed poorly in Cold Diffusion. To address this, they introduced an improved algorithm, formulated as:

\begin{align}
\hat{x}_0 &= R(x_t, t) \label{eq:restorator_operation} \\
x_{t-1} &= x_t - D(\hat{x}_0, t) + D(\hat{x}_0, t-1) \label{eq:new_cd_sampling}
\end{align}

Here, \( \hat{x}_0 \) is the model's predicted output at each step, \( R \) denotes the restoration (denoising) operator (i.e., the DM), and \( D \) represents the degradation (diffusion) process from Equation \ref{eq:cd_noisification}. Unlike conventional DMs that predict noise, this setup directly estimates the original image at each step. Hence, the learning objective for this setup only includes the MSE loss, computed between the predicted and actual images at each timestep. 


\subsubsection{Evaluation}
\label{sec:evaluation_diffusion}
Before testing our approach on downstream tasks, we evaluate the performance of the DMs both qualitatively and quantitatively. We sample 200 images for each model checkpoint throughout training using the same SAR inputs for consistency. The images are sampled using the DDPM method \cite{ho2020}.

For qualitative evaluation, we visually inspect the generated images to evaluate the quality and fidelity of the outputs. For the quantitative evaluation, we compute two standard metrics comparing the generated images to the ground truth: the Structural Similarity Index Measure (SSIM) and the Fréchet Inception Distance (FID). Both metrics aim to evaluate the DMs' ability to generate high-quality and accurate images. More specifically, SSIM measures the structural similarity while FID evaluates the quality and diversity of the generated images. 

\subsection{Downstream Tasks}
After training the DMs, we use it to sample RGB images, which are subsequently used in our selected downstream tasks. We focus on two tasks: (i) land cover classification and (ii) cloud removal. To ensure reliable results, we generate images three times using the same dataset and average the results to reduce randomness. We utilize our DMs at the 250,000th training checkpoint and apply the DDPM sampling method \cite{ho2020} for image generation.

\subsubsection{Land Cover Classification}
Land cover classification involves categorizing images into distinct classes such as vegetation, water bodies, urban areas, agricultural fields, forests, and barren land - an essential task in RS for environmental monitoring and land management.

For this task, we use a supervised learning approach with a pre-trained ResNet50 model from Schmitt \& Wu \cite{schmitt2021imageclassification}. Classification performance is measured using accuracy. We compare the following five input configurations:

\begin{enumerate}
    \item Original S1 (VV channel only)
    \item Original S2 (RGB channels only)
    \item Generated S2 (RGB channels only)
    \item Original S1 \& Original S2 (VV with RGB channels)
    \item Original S1 \& Generated S2 (VV with RGB channels)
\end{enumerate}

Configurations 1, 2, and 4 serve as baselines, while configurations 3 and 5 incorporate images generated during the diffusion process. For multimodal inputs (configurations 4 and 5), we concatenate de modalities on the channel dimension, and modify the ResNet50 model to accept 6-channel input instead of the standard 3-channel input.

\subsubsection{Cloud Removal}
\label{sec:cloud_removal}
Cloud removal is a critical task in RS, as clouds can obstruct essential information in RGB images, hindering accurate analysis and interpretation of satellite data. This is particularly important for applications such as flood detection, where timely and clear imagery is needed.

In our approach, cloud removal is not an explicitly targeted objective but rather an emergent property of our methodology. By leveraging SAR data, which is unaffected by atmospheric conditions, our model inherently learns to reconstruct cloud-free images without directly optimizing for cloud removal. This demonstrates the advantage of using SAR as an auxiliary data source to generate more complete and reliable optical imagery.

To evaluate cloud removal performance, we compare the reconstructed images to the cloud-free Sentinel-2 images - without using any additional downstream model - using three standard metrics: Mean Absolute Error (MAE), Peak Signal-to-Noise Ratio (PSNR), and Structural Similarity Index Measure (SSIM).

\section{Experiments \& Results}
\label{sec:exp_results}

\subsection{Datasets}
We use the following datasets for training our DMs and evaluating downstream tasks. \\

\textbf{SEN12MS:} We train our DMs with the SEN12MS dataset \cite{schmitt2019}, a publicly available dataset containing 180,662 triplets of S1 patches, multi-spectral S2 patches, and MODIS land cover maps. All patches are fully georeferenced at a 10 m ground sampling distance and cover all inhabited continents during all meteorological seasons. To derive the classification labels for each patch, we use the MODIS land cover maps. Specifically, we use the IGBP land cover classes and map them to a simplified classification system as suggested by Schmitt \textit{et al.}\cite{schmitt2021imageclassification}.

%
%

We create the training dataset by randomly sampling 80\% of the full dataset, resulting in $146'000$ samples. From the remaining data, we further sample 200 images to evaluate the performance of our DM. \\

\textbf{DFC 2020:} To evaluate our approach on the land cover classification downstream task, we use the DFC 2020 dataset from the IEEE GRSS 2020 Data Fusion Contest \cite{schmitt2019dfc}. Since this dataset was collected and processed similarly to SEN12-MS, it integrates seamlessly into our pipeline without requiring any additional adaptation or preprocessing. This ensures consistency for direct model evaluation on the downstream task. The dataset comes with a predefined split, from which we use only the test set, containing approximately $5'218$ samples. To train the land-cover classification model, we further divide this test set into an 80/20 split, using 80\% for training and 20\% for evaluation. As with the SEN12-MS dataset, we derive land cover labels for our downstream task from the MODIS segmentation map. \\

\textbf{SEN12MS-CR:} For evaluating our DMs on the cloud removal task, we use the SEN12MS-CR dataset \cite{sen12mscr}. This dataset consists of triplets of S1, cloudy S2, and cloud-free S2 images. Since it follows the same data collection process as SEN12-MS, it also integrates smoothly into our workflow with minimal preprocessing. This dataset includes a predefined split, from which we utilize only the test set containing $7'899$ samples. Since no additional model training is needed for this task, we use the entire test set to report on the metrics.

\subsection{Diffusion Training}

\subsubsection{Implementation Details}
As outlined in Section \ref{sec:model_architecture}, we use the DiT-S architecture with a patch size of $4$. The models process images at their native resolution of $256 \times 256$ pixels. All models are trained using the AdamW optimizer with default parameters and a constant learning rate of $1\mathrm{e}{-4}$, without weight decay. The batch size is set to 192. Each model is trained for 250,000 iterations, which we found sufficient for convergence. Additionally, we employ a linear noise schedule across all experiments, with the number of diffusion steps set to 1000. These settings ensure consistency and reproducibility in our results.

\subsubsection{Experiments}
In this paper, we conduct three experiments to translate SAR images to RGB imagery using DMs. In the first experiment, we train the DM with the Standard Diffusion setup without any class label conditioning, relying solely on the diffusion process to learn the underlying mapping from SAR to RGB. In the second experiment, we introduce class label conditioning to guide the generation process, incorporating land cover labels. In our final experiment, we implement the Cold Diffusion approach inspired by Bansal \textit{et al.} \cite{bansal2022}, as described in Section \ref{sec:cold_diffusion}.

\subsubsection{Evaluation of Diffusion Training}

\begin{figure}[h]
    \centering
    \includegraphics[width=0.5\textwidth]{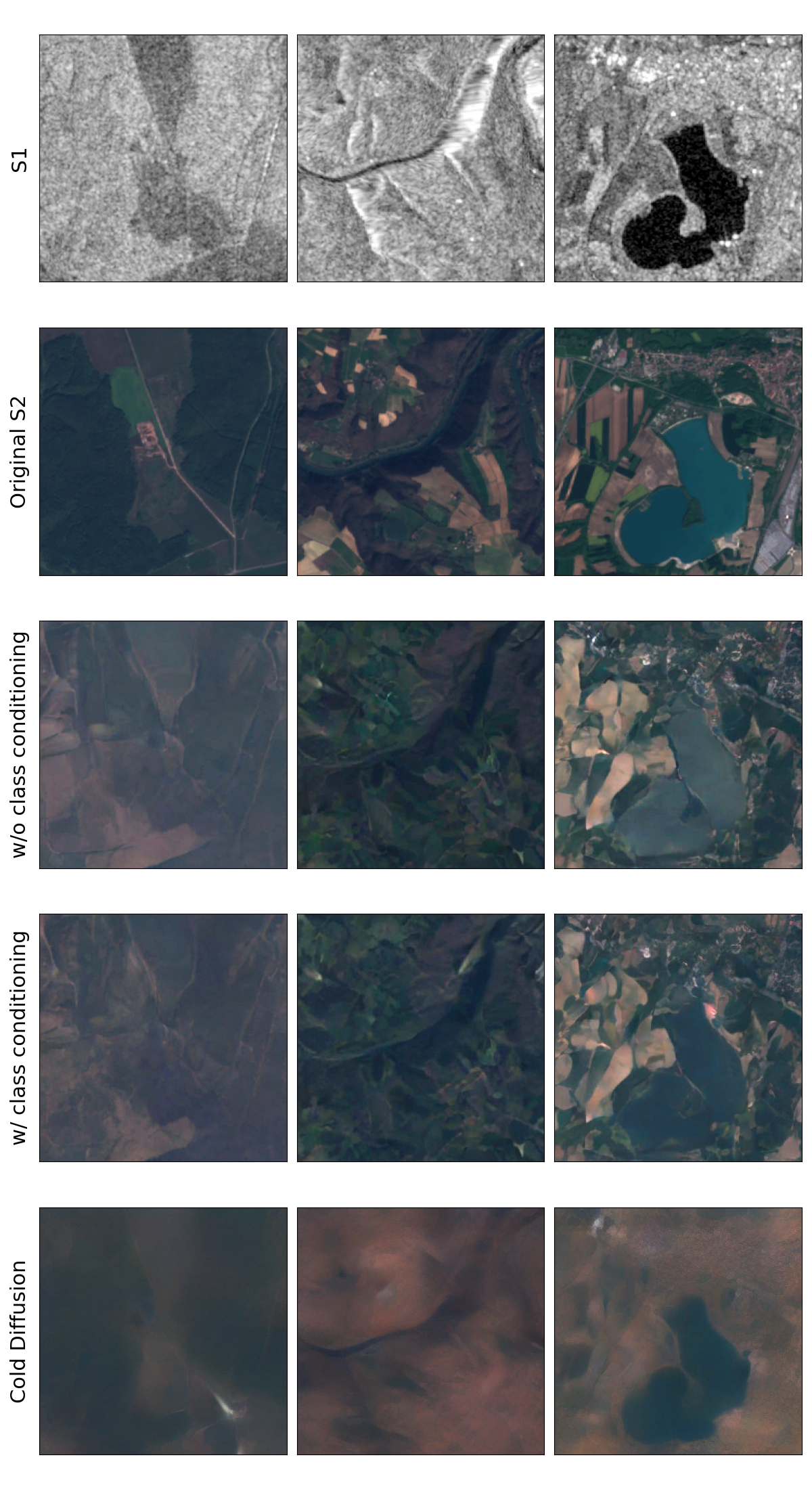}
    \caption{Qualitative comparison of generated S2 images across our experimental setups, presented alongside the original S1 SAR images and corresponding real S2 RGB images. Each column represents a different example, illustrating variations in image translation quality across experiments.}
    \label{fig:generated_images}
\end{figure}

\textbf{Qualitative Evaluation:} Figure \ref{fig:generated_images} presents an overview of the generated RGB images for the Standard Diffusion without class conditioning 1 (3rd row), with class conditioning (4th row) and the Cold Diffusion (5th row) setup - alongside the original SAR and RGB images. Each column represents a different example. 

Comparing the Standard Diffusion experiments without and with class conditioning, the generated samples exhibit only minor differences. Notably, in the images generated with the class conditioning, the colors of the lake in the last column appear darker compared to the original and the generated samples without class labels. This suggests subtle variations in color representation, particularly in water bodies, when class labels are incorporated into the generation process. Despite this, the overall structure and color patterns remain consistent across the samples, indicating a high degree of similarity in the output regardless of the experimental conditions. Our Cold Diffusion experiment displays degradation with over-smooth textures, muted colors, and minimal preservation of spatial details, making the outputs visually less coherent. Overall, we observe that the SAR images are not able to provide sufficient information to perfectly replicate the original RGB images, as there remain notable variances between the generated and original outputs. These discrepancies suggest that while the methods effectively leverage SAR data, the inherent limitations of SAR imagery in capturing certain spectral and contextual details hinder a complete reproduction of the original RGB scenes.

\textbf{Quantitative Evaluation:} To complement the qualitative assessment, we quantitatively evaluate the generated images. Table \ref{tab:quant_eval} presents the quantitative evaluation of our generated images, computed at the 250,000th checkpoint using the DDPM sampling method \cite{ho2020}.

The quantitative evaluation of our DMs is presented in Table \ref{tab:quant_eval}, where we compare SSIM and FID across our experimental setups. The class-conditioned Standard Diffusion experiment slightly outperforms the one without class conditioning, achieving a higher SSIM (0.824 vs. 0.818) and a lower FID (84.681 vs. 85.409), indicating improved structural preservation and visual fidelity. Cold Diffusion achieves the highest SSIM (0.876) but exhibits a significantly higher FID (150.923), suggesting strong structural consistency, but significantly lower perceptual quality as similarly observed in the qualitative evaluation.

We note that the goal of our research is not the visual reconstruction of RGB images but rather to evaluate the performance of these generated images on a specific downstream task. The generated images serve as intermediaries, with their ultimate value determined by their contribution to higher accuracy and robustness in the target application. Consequently, the qualitative evaluation presented here is secondary to the quantitative assessment of task-specific performance. The effectiveness of each setup is best assessed in terms of its impact on downstream tasks rather than standalone image quality metrics.

\begin{table}[htbp]
\centering
\resizebox{0.5\textwidth}{!}{
\begin{tabular}{lcc}
\toprule
\textbf{Setup} & \textbf{SSIM} & \textbf{FID} \\ 
\midrule
Standard Diffusion w/o Class Conditioning & 0.818 & 85.409 \\ 
Standard Diffusion w/ Class Conditioning & 0.824 & \textbf{84.681} \\
Cold Diffusion & \textbf{0.876} & 150.923 \\
\bottomrule
\end{tabular}
}
\caption{Quantitative evaluation of our DMs using Structural Similarity Index (SSIM) and Fréchet Inception Distance (FID). The best results are highlighted in \textbf{bold}.}
\label{tab:quant_eval}
\end{table}


\begin{table*}[htbp]
    \centering
    \begin{tabular}{lccccc}
        \toprule
        \textbf{Setup} & \textbf{S1 only} & \textbf{S2 only} & \textbf{S1 \& S2} & \textbf{Gen S2} & \textbf{S1 \& Gen S2} \\
        \midrule
        \textbf{Baseline} \textit{(real data)} & 0.67 $\pm$ 0.00 & \textbf{0.77 $\pm$ 0.00} & 0.76 $\pm$ 0.00 & - & - \\
        \midrule
        \textbf{Standard Diffusion w/o Class Conditioning} & - & - & - & 0.53 $\pm$ 0.00 & 0.69 $\pm$ 0.02 \\
        \textbf{Standard Diffusion w/ Class Conditioning}  & - & - & - & \textbf{0.69 $\pm$ 0.01} & \textbf{0.73 $\pm$ 0.01} \\
        \textbf{Cold Diffusion} & - & - & - & 0.60 $\pm$ 0.01 & 0.69 $\pm$ 0.02 \\
        \bottomrule
    \end{tabular}
    \caption{Quantitative evaluation of land cover classification performance for our experimental setups. Results are reported as mean $\pm$ standard deviation. The evaluation considers different input configurations: original S1 (S1 only), original S2 (S2 only), both original S1 and S2 (S1 \& S2), generated S2 (Gen S2), and the combination of original S1 with generated S2 (S1 \& Gen S2). The best results for the experimental setups and for the baseline are highlighted in \textbf{bold}.}
    \label{tab:res_classification}
\end{table*}



\subsection{Downstream Applications}
\subsubsection{Land Cover Classification}
In this section, we present the results of each experiment, analyzing their impact on land cover classification performance. We evaluate the effectiveness of the generated images in this downstream task and compare the classification outcomes to the baseline. Each experiment is trained for 20 epochs using the AdamW optimizer with its default hyperparameters. The learning rate and weight decay are both set to $5\mathrm{e}{-5}$ and $1\mathrm{e}{-4}$, respectively. The batch size remains constant at 10 throughout all experiments.

Table \ref{tab:res_classification} presents the classification performance across different experimental setups, comparing S1, S2, and generated S2 data. The baseline results indicate that using only RGB imagery achieves the highest accuracy ($0.77$), while using only SAR data results in a lower accuracy of $0.67$. Combining both SAR and RGB reduces the performance over using RGB alone, yielding an accuracy of $0.76$. This suggests that the additional SAR data does not significantly enhance classification when the original RGB imagery is already available. These results align with expectations, as RGB imagery typically provides more direct information about surface characteristics relevant for classification, while SAR data captures structural and moisture-related features that may be less discriminative for certain land cover types.

When evaluating the experiments that incorporate generated RGB images, it is evident that synthetic RGB data introduces additional challenges. In the Standard Diffusion setup without class conditioning, using only generated RGB images results in a significantly lower accuracy of $0.53$, indicating that the generated images lack some critical information present in real RGB imagery. However, when combining S1 with Gen S2, the accuracy improves to $0.69$, slightly higher than the S1 only setup ($0.67$). This suggests that the synthetic optical data still provides some useful information when paired with SAR data. In the Standard Diffusion setup with class conditioning, the classification accuracy improves across the board. Notably, using generated RGB images alone achieves a much higher accuracy of $0.69$ compared to the first setup, suggesting that class conditioning improves the quality and relevance of the generated RGB images. The inclusion of the original SAR images also benefits from class conditioning, reaching $0.73$, though it still falls short of both the S2 only setup ($0.77$) and original S1 \& S2 combination ($0.76$). The Cold Diffusion setup achieves moderate improvements over the non-class conditioned Standard Diffusion setup with an accuracy of $0.60$ for the generated RGB images alone and $0.69$ when combined with SAR images. However, it underperformed compared to our class-conditioned model and the baseline. 

Comparing across our experimental setups, class conditioning proves to be a valuable technique for improving the performance of models that rely on generated optical data. The improvements in Gen S2-based classification suggest that the conditioning mechanism helps guide the generation process to produce more class-discriminative images. Notably, our findings indicate that standard quantitative and qualitative evaluations do not always correlate with downstream task performance. For instance, while Cold Diffusion-generated images exhibited lower visual fidelity and higher visual degradation, they still outperformed the Standard Diffusion setup without class conditioning in classification tasks. Similarly, a high degree of structural integrity does not necessarily translate to improved downstream performance, as the class-conditioned setup surpassed the Cold Diffusion setup.

Overall, our results indicate that synthetic RGB data does not fully replace real RGB imagery yet, as performance remains lower than the baseline S2 results. Nevertheless, these findings highlight the potential of generative approaches to enhance SAR-based classification in scenarios where RGB data is missing or unavailable.

\subsubsection{Cloud Removal}
In this section, we present the results of our cloud removal experiments, evaluating the ability of our generative model to reconstruct clear S2 images from cloud-covered inputs. As we utilize the \textbf{Sen12MS-CR} dataset for this task, our experimental design is constrained by the dataset's limitations. Specifically, the dataset does not provide land cover labels, restricting our evaluation to our Standard Diffusion without class conditioning and Cold Diffusion. 

Table \ref{tab:res_cloudremoval} presents the quantitative evaluation results for cloud removal performance, comparing various state-of-the-art methods against our Standard Diffusion without class conditioning and Cold Diffusion setup. Our Standard Diffusion setup without class conditioning achieves a MAE of 0.023 and SSIM of 0.816, whereas the Cold Diffusion setup achieves $0.020$ and $0.871$, respectively. Both approaches outperformed most GAN-based approaches and produced results that are slightly lower than the best-performing cloud removal methods such as \textit{DiffCR} and \textit{UnCRtainTS L2}. However, it is important to note that while these methodologies are specifically designed for cloud removal, our model is not explicitly optimized for this task. Despite this, our setups demonstrate competitive performance, particularly in PSNR ($31.85$ and $32.84$), indicating that it effectively preserves image quality. This highlights the versatility and potential of our approach in handling complex image translation tasks, even in cases where they are not directly tailored to the given problem.

\begin{table}[htbp]
    \centering
    \resizebox{0.5\textwidth}{!}{
    \begin{tabular}{lccc}
        \toprule
        \textbf{Method} & \textbf{MAE} & \textbf{PSNR} & \textbf{SSIM} \\
        \midrule
        McGAN \cite{mcgan}                   & $0.048$ & $25.14$ & $0.744$ \\
        SAR-Opt-cGAN \cite{saroptcgan}            & $0.043$ & $25.59$ & $0.764$ \\
        SAR2OPT \cite{sar2opt}                 & $0.042$ & $25.87$ & $0.793$ \\
        SpA GAN \cite{spagan}                 & $0.045$ & $24.78$ & $0.754$ \\
        Simulation-Fusion GAN \cite{simulationfusiongan}   & $0.045$ & $24.73$ & $0.701$ \\
        DSen2-CR \cite{dsen2cr}                & $0.031$ & $27.76$ & $0.874$ \\
        GLF-CR \cite{glfcr}                  & $0.028$ & $28.64$ & $0.885$ \\
        UnCRtainTS L2 \cite{uncrtaintsl2}           & $0.027$ & $28.90$ & $0.880$ \\
        DiffCR \cite{zou2023}                  & $\textbf{0.019}$ & $31.77$ & $\textbf{0.902}$ \\
        \rowcolor{gray!20} \textbf{Standard Diffusion w/o Class Conditioning}             & $0.023$ & $31.85$ & $0.816$ \\
        \rowcolor{gray!20} \textbf{Cold Diffusion} & $0.020$ & $\textbf{32.84}$ & $0.871$ \\
        \bottomrule
    \end{tabular}
    }
    \caption{Comparison of Cloud Removal Performance between the Standard Diffusion setup without class conditioning, Cold Diffusion and other existing methods. Results are reported as mean values, with the best scores for each metric highlighted in \textbf{bold}. The rows highlighted in grey represent our results.} 
    \label{tab:res_cloudremoval}
\end{table}
\section{Conclusion}
\label{sec:conclusion}

In this work, we introduce a novel approach for translating SAR images to RGB images using a ViT-based Diffusion Model (DM). Our methodology addresses the challenge of missing or obstructed optical imagery by leveraging a three-step process: first, training a DM to learn the mapping from SAR to RGB; second, generating synthetic S2 images from S1 inputs; and third, systematically assessing the practical usability of the generated images in downstream RS applications, specifically land cover classification and cloud removal.

Our experimental results reveal that while generated images may not fully replicate real S2 data, they still contribute valuable information in classification and image restoration tasks. Specifically, our class-conditioned Standard Diffusion model achieves higher classification accuracy, demonstrating the benefit of incorporating land cover labels during training. Additionally, despite exhibiting lower perceptual quality, the Cold Diffusion setup performs competitively in land cover classification, indicating that conventional quantitative evaluation metrics may not fully capture the effectiveness of generated images in downstream tasks. Our cloud removal experiments show that our approach produces competitive results against state-of-the-art methods, despite not being explicitly optimized for this task. These findings suggest that diffusion-based SAR-to-optical translation has the potential to enhance RS applications where S2 images are incomplete or missing.
Despite these promising results, several challenges remain to be addressed. The generated images exhibit limitations in capturing fine-grained spectral details, which can impact their effectiveness in certain applications. Future research could explore more advanced fusion techniques for SAR conditioning, such as incorporating attention mechanisms instead of simple concatenation to better capture the global context and multi-scale features of SAR imagery. Additionally, integrating an adversarial loss for land cover classification could enhance the utility of the generated images. Further improvements may also be achieved by incorporating additional spectral bands or leveraging specialized architectures designed to address SAR-specific challenges, such as mitigating speckle noise and modeling the complex relationship between radar backscatter and visible features.

{
    \small
    \bibliographystyle{ieeenat_fullname}
    \bibliography{main}

\begin{thebibliography}{40}
\providecommand{\natexlab}[1]{#1}
\providecommand{\url}[1]{\texttt{#1}}
\expandafter\ifx\csname urlstyle\endcsname\relax
  \providecommand{\doi}[1]{doi: #1}\else
  \providecommand{\doi}{doi: \begingroup \urlstyle{rm}\Url}\fi

\bibitem[Adriano et~al.(2020)Adriano, Yokoya, Xia, Miura, Liu, Matsuoka, and Koshimura]{Adriano2020LearningFM}
Bruno Adriano, Naoto Yokoya, Junshi Xia, Hiroyuki Miura, Wen Liu, Masashi Matsuoka, and Shunichi Koshimura.
\newblock Learning from multimodal and multitemporal earth observation data for building damage mapping.
\newblock \emph{ArXiv}, abs/2009.06200, 2020.

\bibitem[AI(2022)]{stabilityai_sd_vae}
Stability AI.
\newblock Stable diffusion vae - fine-tuned with ema, 2022.
\newblock Accessed: 2025-01-23.

\bibitem[Bai et~al.(2024)Bai, Pu, and Xu]{bai2024_first}
Xinyu Bai, Xinyang Pu, and Feng Xu.
\newblock Conditional diffusion for sar to optical image translation.
\newblock \emph{IEEE Geoscience and Remote Sensing Letters}, 21:\penalty0 1--5, 2024.

\bibitem[Bansal et~al.(2022)Bansal, Borgnia, Chu, Li, Kazemi, Huang, Goldblum, Geiping, and Goldstein]{bansal2022}
Arpit Bansal, Eitan Borgnia, Hong-Min Chu, Jie~S. Li, Hamid Kazemi, Furong Huang, Micah Goldblum, Jonas Geiping, and Tom Goldstein.
\newblock Cold diffusion: Inverting arbitrary image transforms without noise, 2022.

\bibitem[Berian et~al.(2025)Berian, Brignac, Wu, Daba, and Mahalanobis]{berian2025}
Alex Berian, Daniel Brignac, JhihYang Wu, Natnael Daba, and Abhijit Mahalanobis.
\newblock Crossmodalitydiffusion: Multi-modal novel view synthesis with unified intermediate representation, 2025.

\bibitem[Bermudez et~al.(2018)Bermudez, Happ, Oliveira, and Feitosa]{sar2opt}
J.~D. Bermudez, P.~N. Happ, D.~A.~B. Oliveira, and R.~Q. Feitosa.
\newblock Sar to optical image synthesis for cloud removal with generative adversarial networks.
\newblock \emph{ISPRS Annals of the Photogrammetry, Remote Sensing and Spatial Information Sciences}, IV-1:\penalty0 5--11, 2018.

\bibitem[Chen et~al.(2022)Chen, Zhang, Li, Wang, and Zhang]{dsen2cr}
Shanjing Chen, Wenjuan Zhang, Zhen Li, Yuxi Wang, and Bing Zhang.
\newblock Cloud removal with sar-optical data fusion and graph-based feature aggregation network.
\newblock \emph{Remote Sensing}, 14\penalty0 (14), 2022.

\bibitem[Dhariwal and Nichol(2021)]{Dhariwal2021DiffusionMB}
Prafulla Dhariwal and Alex Nichol.
\newblock Diffusion models beat gans on image synthesis.
\newblock \emph{ArXiv}, abs/2105.05233, 2021.

\bibitem[Dosovitskiy et~al.(2021)Dosovitskiy, Beyer, Kolesnikov, Weissenborn, Zhai, Unterthiner, Dehghani, Minderer, Heigold, Gelly, Uszkoreit, and Houlsby]{dosovitskiy2021}
Alexey Dosovitskiy, Lucas Beyer, Alexander Kolesnikov, Dirk Weissenborn, Xiaohua Zhai, Thomas Unterthiner, Mostafa Dehghani, Matthias Minderer, Georg Heigold, Sylvain Gelly, Jakob Uszkoreit, and Neil Houlsby.
\newblock An image is worth 16x16 words: Transformers for image recognition at scale, 2021.

\bibitem[Drusch et~al.(2012)Drusch, Del~Bello, Carlier, Colin, Fernandez, Gascon, Hoersch, Isola, Laberinti, Martimort, et~al.]{drusch2012sentinel}
Matthias Drusch, Umberto Del~Bello, S{\'e}bastien Carlier, Olivier Colin, Veronica Fernandez, Ferran Gascon, Bianca Hoersch, Claudia Isola, Paolo Laberinti, Philippe Martimort, et~al.
\newblock Sentinel-2: Esa's optical high-resolution mission for gmes operational services.
\newblock \emph{Remote sensing of Environment}, 120:\penalty0 25--36, 2012.

\bibitem[Ebel et~al.(2020)Ebel, Meraner, Schmitt, and Zhu]{sen12mscr}
Patrick Ebel, Andrea Meraner, Michael Schmitt, and Xiao~Xiang Zhu.
\newblock {Multisensor Data Fusion for Cloud Removal in Global and All-Season Sentinel-2 Imagery}.
\newblock \emph{IEEE Transactions on Geoscience and Remote Sensing}, 2020.

\bibitem[Ebel et~al.(2023)Ebel, Garnot, Schmitt, Wegner, and Zhu]{uncrtaintsl2}
Patrick Ebel, Vivien Sainte~Fare Garnot, Michael Schmitt, Jan~Dirk Wegner, and Xiao~Xiang Zhu.
\newblock Uncrtaints: Uncertainty quantification for cloud removal in optical satellite time series, 2023.

\bibitem[Enomoto et~al.(2017)Enomoto, Sakurada, Wang, Fukui, Matsuoka, Nakamura, and Kawaguchi]{mcgan}
Kenji Enomoto, Ken Sakurada, Weimin Wang, Hiroshi Fukui, Masashi Matsuoka, Ryosuke Nakamura, and Nobuo Kawaguchi.
\newblock Filmy cloud removal on satellite imagery with multispectral conditional generative adversarial nets.
\newblock In \emph{2017 IEEE Conference on Computer Vision and Pattern Recognition Workshops (CVPRW)}, pages 1533--1541, 2017.

\bibitem[Fu et~al.(2019)Fu, Xu, and Jin]{fu2019}
Shilei Fu, Feng Xu, and Ya-Qiu Jin.
\newblock Reciprocal translation between sar and optical remote sensing images with cascaded-residual adversarial networks, 2019.

\bibitem[Gao et~al.(2020)Gao, Yuan, Li, Zhang, and Su]{simulationfusiongan}
Jianhao Gao, Qiangqiang Yuan, Jie Li, Hai Zhang, and Xin Su.
\newblock Cloud removal with fusion of high resolution optical and sar images using generative adversarial networks.
\newblock \emph{Remote Sensing}, 12\penalty0 (1), 2020.

\bibitem[Goodfellow et~al.(2014)Goodfellow, Pouget-Abadie, Mirza, Xu, Warde-Farley, Ozair, Courville, and Bengio]{Goodfellow2014GenerativeAN}
Ian~J. Goodfellow, Jean Pouget-Abadie, Mehdi Mirza, Bing Xu, David Warde-Farley, Sherjil Ozair, Aaron~C. Courville, and Yoshua Bengio.
\newblock Generative adversarial networks.
\newblock \emph{Communications of the ACM}, 63:\penalty0 139 -- 144, 2014.

\bibitem[Grohnfeldt et~al.(2018)Grohnfeldt, Schmitt, and Zhu]{saroptcgan}
Claas Grohnfeldt, Michael Schmitt, and Xiaoxiang Zhu.
\newblock A conditional generative adversarial network to fuse sar and multispectral optical data for cloud removal from sentinel-2 images.
\newblock In \emph{IGARSS 2018 - 2018 IEEE International Geoscience and Remote Sensing Symposium}, pages 1726--1729, 2018.

\bibitem[Hafner et~al.(2022)Hafner, Nascetti, Azizpour, and Ban]{haftner2022}
Sebastian Hafner, Andrea Nascetti, Hossein Azizpour, and Yifang Ban.
\newblock Sentinel-1 and sentinel-2 data fusion for urban change detection using a dual stream u-net.
\newblock \emph{IEEE Geoscience and Remote Sensing Letters}, 19:\penalty0 1--5, 2022.

\bibitem[Han et~al.(2023)Han, Zhao, Lv, Zhang, Liu, Bi, and Han]{han2023}
Lintao Han, Yuchen Zhao, Hengyi Lv, Yisa Zhang, Hailong Liu, Guoling Bi, and Qing Han.
\newblock Enhancing remote sensing image super-resolution with efficient hybrid conditional diffusion model.
\newblock \emph{Remote Sensing}, 15\penalty0 (13), 2023.

\bibitem[Hanna et~al.(2023{\natexlab{a}})Hanna, Borth, and Mommert]{Hanna2023PhysicsGuidedML}
Joelle Hanna, Damian Borth, and Michael Mommert.
\newblock Physics-guided multitask learning for estimating power generation and co2 emissions from satellite imagery.
\newblock \emph{IEEE Transactions on Geoscience and Remote Sensing}, 61:\penalty0 1--12, 2023{\natexlab{a}}.

\bibitem[Hanna et~al.(2023{\natexlab{b}})Hanna, Mommert, and Borth]{Hanna2023SparseMV}
Joelle Hanna, Michael Mommert, and Damian Borth.
\newblock Sparse multimodal vision transformer for weakly supervised semantic segmentation.
\newblock \emph{2023 IEEE/CVF Conference on Computer Vision and Pattern Recognition Workshops (CVPRW)}, pages 2145--2154, 2023{\natexlab{b}}.

\bibitem[Ho and Salimans(2022)]{ho2022}
Jonathan Ho and Tim Salimans.
\newblock Classifier-free diffusion guidance, 2022.

\bibitem[Ho et~al.(2020)Ho, Jain, and Abbeel]{ho2020}
Jonathan Ho, Ajay Jain, and Pieter Abbeel.
\newblock Denoising diffusion probabilistic models, 2020.

\bibitem[Hänsch(2019)]{schmitt2019dfc}
Michael Schmitt; Lloyd Hughes; Pedram Ghamisi; Naoto Yokoya;~Ronny Hänsch.
\newblock 2020 ieee grss data fusion contest, 2019.

\bibitem[Khanna et~al.(2024)Khanna, Liu, Zhou, Meng, Rombach, Burke, Lobell, and Ermon]{khanna2024}
Samar Khanna, Patrick Liu, Linqi Zhou, Chenlin Meng, Robin Rombach, Marshall Burke, David Lobell, and Stefano Ermon.
\newblock Diffusionsat: A generative foundation model for satellite imagery, 2024.

\bibitem[Liu et~al.(2024)Liu, Yue, Xia, Ghamisi, Xie, and Fang]{Liu_2024}
Yidan Liu, Jun Yue, Shaobo Xia, Pedram Ghamisi, Weiying Xie, and Leyuan Fang.
\newblock Diffusion models meet remote sensing: Principles, methods, and perspectives.
\newblock \emph{IEEE Transactions on Geoscience and Remote Sensing}, 62:\penalty0 1–22, 2024.

\bibitem[Notarnicola et~al.(2017)Notarnicola, Asam, Jacob, Marin, Rossi, and Stendardi]{Notarnicola2017}
C. Notarnicola, S. Asam, A. Jacob, C. Marin, M. Rossi, and L. Stendardi.
\newblock Mountain crop monitoring with multitemporal sentinel-1 and sentinel-2 imagery.
\newblock In \emph{2017 9th International Workshop on the Analysis of Multitemporal Remote Sensing Images (MultiTemp)}, pages 1--4, 2017.

\bibitem[Pan(2020)]{spagan}
Heng Pan.
\newblock Cloud removal for remote sensing imagery via spatial attention generative adversarial network, 2020.

\bibitem[Peebles and Xie(2023)]{peebles2023}
William Peebles and Saining Xie.
\newblock Scalable diffusion models with transformers, 2023.

\bibitem[Prodromou et~al.(2023)Prodromou, Theocharidis, Fotiou, Argyriou, Polydorou, Alatza, Pittaki, Hadjimitsis, and Tzouvaras]{podromou2023}
Maria Prodromou, Christos Theocharidis, Kyriaki Fotiou, Athanasios~V. Argyriou, Thomaida Polydorou, Stavroula Alatza, Zampela Pittaki, Diofantos Hadjimitsis, and Marios Tzouvaras.
\newblock Rapid landslide mapping using multi-temporal image composites from sentinel-1 and sentinel-2 imagery through google earth engine.
\newblock In \emph{IGARSS 2023 - 2023 IEEE International Geoscience and Remote Sensing Symposium}, pages 2596--2599, 2023.

\bibitem[Ramos and Sappa(2024)]{ThomasRamos2024MultispectralSS}
Leo~Thomas Ramos and Angel~Domingo Sappa.
\newblock Multispectral semantic segmentation for land cover classification: An overview.
\newblock \emph{IEEE Journal of Selected Topics in Applied Earth Observations and Remote Sensing}, 17:\penalty0 14295--14336, 2024.

\bibitem[Rombach et~al.(2022)Rombach, Blattmann, Lorenz, Esser, and Ommer]{rombach2022}
Robin Rombach, Andreas Blattmann, Dominik Lorenz, Patrick Esser, and Björn Ommer.
\newblock High-resolution image synthesis with latent diffusion models, 2022.

\bibitem[Ronneberger et~al.(2015)Ronneberger, Fischer, and Brox]{Ronneberger2015UNetCN}
Olaf Ronneberger, Philipp Fischer, and Thomas Brox.
\newblock U-net: Convolutional networks for biomedical image segmentation.
\newblock \emph{ArXiv}, abs/1505.04597, 2015.

\bibitem[Scheibenreif et~al.(2022)Scheibenreif, Hanna, Mommert, and Borth]{Scheibenreif2022SelfsupervisedVT}
Linus Scheibenreif, Joelle Hanna, Michael Mommert, and Damian Borth.
\newblock Self-supervised vision transformers for land-cover segmentation and classification.
\newblock \emph{2022 IEEE/CVF Conference on Computer Vision and Pattern Recognition Workshops (CVPRW)}, pages 1421--1430, 2022.

\bibitem[Schmitt and Wu(2021)]{schmitt2021imageclassification}
Michael Schmitt and Yu-Lun Wu.
\newblock Remote sensing image classification with the sen12ms dataset, 2021.

\bibitem[Schmitt et~al.(2019)Schmitt, Hughes, Qiu, and Zhu]{schmitt2019}
Michael Schmitt, Lloyd~Haydn Hughes, Chunping Qiu, and Xiao~Xiang Zhu.
\newblock Sen12ms -- a curated dataset of georeferenced multi-spectral sentinel-1/2 imagery for deep learning and data fusion, 2019.

\bibitem[Sohl-Dickstein et~al.(2015)Sohl-Dickstein, Weiss, Maheswaranathan, and Ganguli]{sohl2015}
Jascha Sohl-Dickstein, Eric~A. Weiss, Niru Maheswaranathan, and Surya Ganguli.
\newblock Deep unsupervised learning using nonequilibrium thermodynamics.
\newblock In \emph{International Conference on Machine Learning}, pages 2256--2265. PMLR, 2015.

\bibitem[Torres et~al.(2012)Torres, Snoeij, Geudtner, Bibby, Davidson, Attema, Potin, Rommen, Floury, Brown, et~al.]{torres2012gmes}
Ramon Torres, Paul Snoeij, Dirk Geudtner, David Bibby, Malcolm Davidson, Evert Attema, Pierre Potin, Bj{\"O}rn Rommen, Nicolas Floury, Mike Brown, et~al.
\newblock Gmes sentinel-1 mission.
\newblock \emph{Remote sensing of environment}, 120:\penalty0 9--24, 2012.

\bibitem[Xu et~al.(2022)Xu, Shi, Ebel, Yu, Xia, Yang, and Zhu]{glfcr}
Fang Xu, Yilei Shi, Patrick Ebel, Lei Yu, Gui-Song Xia, Wen Yang, and Xiao~Xiang Zhu.
\newblock Glf-cr: Sar-enhanced cloud removal with global-local fusion, 2022.

\bibitem[Zou et~al.(2023)Zou, Li, Xing, Zhang, Wang, Jin, and Tao]{zou2023}
Xuechao Zou, Kai Li, Junliang Xing, Yu Zhang, Shiying Wang, Lei Jin, and Pin Tao.
\newblock Diffcr: A fast conditional diffusion framework for cloud removal from optical satellite images, 2023.

\end{thebibliography}
}


\end{document}